\pdfoutput=1

\documentclass[11pt]{article}

\usepackage[]{EMNLP2023}

\usepackage{times}
\usepackage{latexsym}

\usepackage[T1]{fontenc}

\usepackage[utf8]{inputenc}

\usepackage{microtype}

\usepackage{inconsolata}

\usepackage{microtype}
\usepackage{graphicx}
\usepackage{multirow}
\usepackage{multicol}
\usepackage{microtype}
\usepackage{subcaption}
\usepackage{enumitem,kantlipsum}
\usepackage[normalem]{ulem}
\usepackage{tabularx}
\usepackage{tcolorbox}
\usepackage{color, colortbl}
\usepackage{todonotes} 
\usepackage{amsmath}
\usepackage{pbox}
\usepackage{booktabs} 
\usepackage{amssymb}
\usepackage{pifont}

\usepackage{tabu}
\usepackage{fontawesome}

\title{Nexus at ArAIEval Shared Task: 
Fine-Tuning Arabic Language Models for Propaganda and Disinformation Detection
}

\author{Yunze Xiao$^1$, Firoj Alam$^2$\\
  $^1$Carnegie Mellon University in Qatar, Doha, Qatar\\
  $^2$Qatar Computing Research Institute, HBKU, Doha, Qatar \\
  \texttt{yunzex@andrew.cmu.edu,falam@hbku.edu.qa},
  \\}

\begin{document}
\maketitle
\begin{abstract}

The spread of disinformation and propagandistic content poses a threat to societal harmony, undermining informed decision-making and trust in reliable sources. Online platforms often serve as breeding grounds for such content, and malicious actors exploit the vulnerabilities of audiences to shape public opinion. Although there have been research efforts aimed at the automatic identification of disinformation and propaganda in social media content, there remain challenges in terms of performance. The ArAIEval shared task aims to further research on these particular issues within the context of the Arabic language. In this paper, we discuss our participation in these shared tasks. We competed in subtasks 1A and 2A, where our submitted system secured positions 9th and 10th, respectively. Our experiments consist of fine-tuning transformer models and using zero- and few-shot learning with GPT-4.

\end{abstract}

\section{Introduction}
\label{sec:introduction}

In various communication channels, propaganda, also known as persuasive techniques, is disseminated through a wide set of methods. These techniques can range from appealing to the audience's emotions—known as the \textit{``emotional technique''} — to employing logical fallacies. Examples of such fallacies include \textit{``straw man''} arguments, which misrepresent someone's opinion; covert \textit{``ad hominem''} attacks; and \textit{``red herrings''}, which introduce irrelevant data to divert attention from the issue at hand~\cite{Miller}.

Previous research in this area has taken various approaches to identify propagandistic content. These include assessing content based on writing style and readability levels in articles~\cite{rashkin-EtAl:2017:EMNLP2017,BARRONCEDENO20191849}, examining sentences and specific fragments within news articles using fine-grained techniques~\cite{EMNLP19DaSanMartino}, as well as evaluating memes for propagandistic elements~\cite{dimitrov2021detecting}. 

Moreover, malicious actors manipulate media platforms to shape public opinion, disseminate hate speech, target individuals' subconscious minds, spread offensive content, and fabricate falsehoods, among other. These efforts are part of broader strategies to influence people's thoughts and actions~\cite{Zhou2016,alam-etal-2022-survey,ijcai2022p781}.

\begin{figure}
    \centering
    \includegraphics[width=0.8\linewidth]{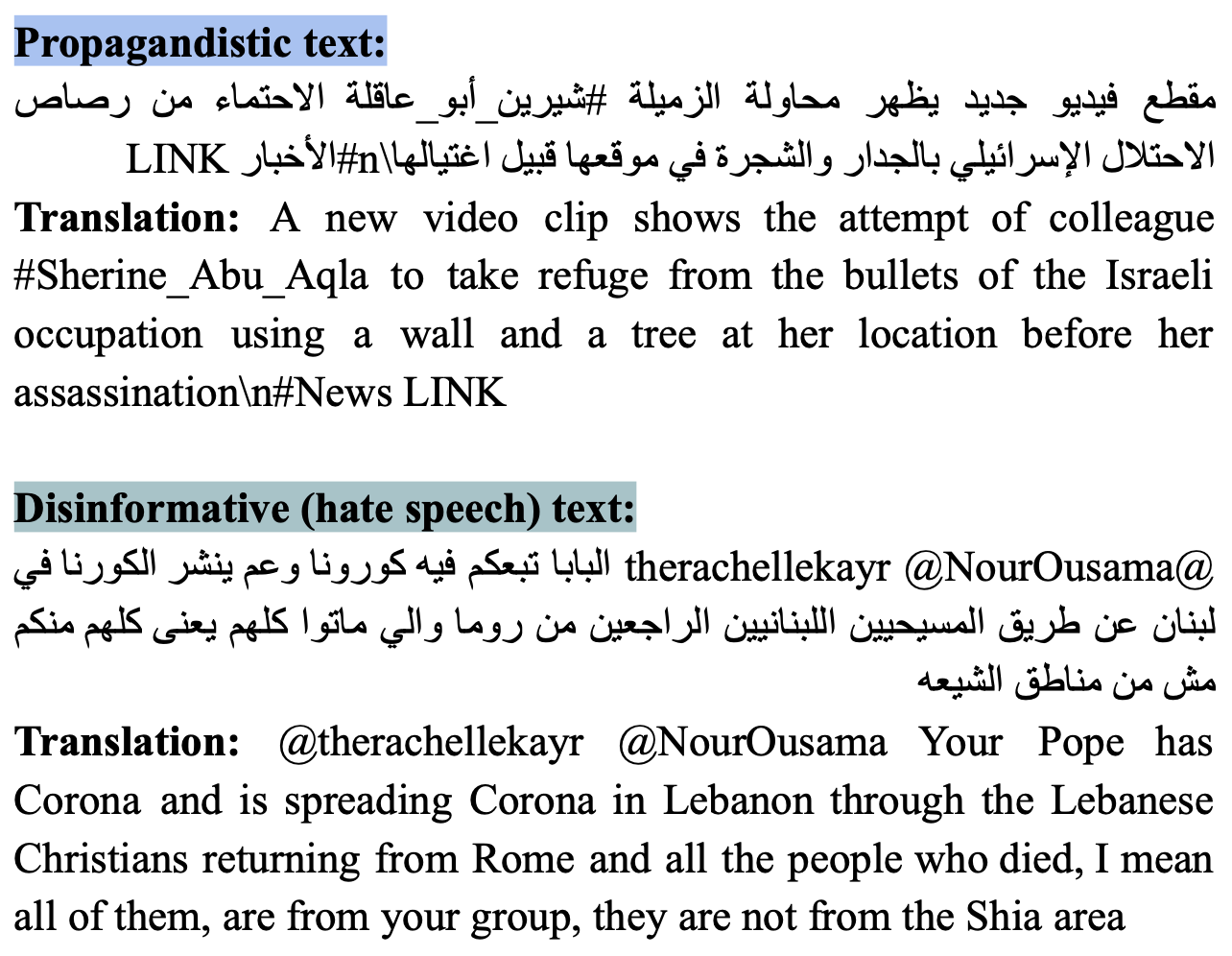}
    \caption{Examples of propagandistic and disinformative text.}
    \label{fig:examples}
\end{figure}

In a broader context, the proliferation of such disinformation can pose significant threats to societal harmony and undermine the trust individuals have in reliable sources \cite{mubarak2023detecting}. Currently, these manipulative strategies are widespread across various online platforms, where they are employed to influence public opinion and distort perceptions, taking advantage of the vulnerabilities of unsuspecting audiences \cite{oshikawa2018survey, oshikawa-etal-2020-survey}.

The far-reaching consequences of misinformation and propaganda include the incitement of prejudices and discriminatory behaviors, as well as the exacerbation of social divisions and polarization \cite{fortuna2018survey,Offensive:Type:Target:2019,zampieri2020semeval,EMNLP19DaSanMartino}. In extreme cases, such false narratives can even fuel radicalization, threatening societal stability. Ultimately, the spread of misinformation undermines democracy by depriving citizens of the accurate information needed for informed decision-making \cite{li2016survey}. The digital age has expanded the reach of propaganda, subtly influencing individuals' perspectives even in their most private spheres.

Since propaganda can manifest in a variety of forms, detecting it and other types of misinformation has always been a challenging task. This task necessitates a deeper analysis of the context in which the content is presented. Therefore, the goal of the shared task is to advance research by developing methods and algorithms for identifying disinformation and propagandistic content. In Figure \ref{fig:examples}, we provide examples that depict such content.

In the ArAIEval shared task at ArabicNLP 2023~\cite{araieval:arabicnlp2023-overview}, there are two tasks with two subtasks each: {\em(i)} \textbf{Task 1 Persuasion Technique Detection} and {\em(ii)} \textbf{Task 2: Disinformation Detection}. Each has two subtasks. We used pre-trained transformer-based models to fine-tune them on the task specific datasets.

We participated in subtasks 1A and 2A, where we fine-tuned pretrained models to predict whether the texts contain persuasion techniques (1A) or are disinformative (2A). We also explored zero-shot and few-shot learning using GPT-4 to understand its performance for these tasks. Both subtasks in which we participated fall under binary classification settings.

\section{Related Work}
\label{sec:related_work}

In this section, we discuss the research related to the automatic detection of persuasion techniques and disinformation. 

Over the past few decades, the use of persuasion techniques, often in the form of propaganda, has proliferated on social media platforms, aiming to influence or mislead audiences. This has become a major concern for a wide range of stakeholders, including social media companies and government agencies. In response to this growing issue, the emerging field of "computational propaganda" aims to automatically identify such manipulative techniques across various forms of content—textual, visual, and multimodal (e.g., memes). 

Recently, the study by \cite{EMNLP19DaSanMartino} curated a variety of persuasive techniques. These range from emotional manipulations, such as using \textit{Loaded Language} and \textit{Appeal to Fear}, to logical fallacies like \textit{Straw Man} (misrepresenting someone's opinion) and \textit{Red Herring} (introducing irrelevant data). The study primarily focused on textual content, such as newspaper articles. In a similar vein, \cite{da2020semeval} organized a shared task on the "Detection of Propaganda Techniques in News Articles." Building on these previous efforts, \cite{SemEval2021-6-Dimitrov}\footnote{\url{http://propaganda.math.unipd.it/semeval2021task6/}} orchestrated the \textit{SemEval-2021 Shared Task 6 on Detection of Propaganda Techniques in Memes} in 2021. This task had a multimodal setup, integrating both text and images, and challenged participants to construct systems capable of identifying the propaganda techniques employed in specific memes. Efforts have also been made towards multilingual propaganda detection. \cite{hasanain-etal-2023-qcri} demonstrates that multilingual models significantly outperform monolingual ones, even in languages that are unseen.


While most of these efforts have focused primarily on English, \citet{propaganda-detection:WANLP2022-overview} organized a shared task on fine-grained propaganda techniques in Arabic to enrich the field of Arabic AI research. This event attracted numerous participants. 

In addition to the use of propaganda, malicious social media users frequently disseminate disinformative content—including hate speech, offensive material, rumors, and spam—to advance social and political agendas or to harm individuals, entities, and organizations. To address this issue, the current literature has explored automated techniques for detecting disinformation on social media platforms. For example, the study by \citet{Demilie22} investigated the detection of fake news and hate speech in Ethiopian social media. The researchers found that a hybrid approach, combining both deep learning and traditional machine learning techniques, proved to be the most effective in identifying disinformation in that context. 

In the field of Arabic social media, numerous researchers have used various approaches for disinformation detection. For example, the study by \citet{Boulouard22} focused on identifying hate speech and offensive content in Arabic social media platforms. By employing transfer learning techniques, they found that BERT~\cite{devlin2018bert} and AraBERT~\cite{baly2020arabert} yielded the highest accuracy rates, at 98\% and 96\%, respectively. Other significant contributions to the area of Arabic hate speech and offensive content detection include works by \citet{zampieri2020semeval} and \citet{mubarak2020overview}.

\section{Task and Dataset}
\label{sec:data}


As discussed earlier we used the datasets released as a part of the ArAIEval shared task \cite{araieval:arabicnlp2023-overview}. We participated in subtask 1A and 2A. They are defined as follows. 
\paragraph{Subtask 1A:}  Given a multigenre (tweet and news paragraphs of the news articles) snippet, identify whether it contains content with persuasion technique. This is a binary classification task.

The data for Subtask 1A is composed of IDs, text, and labels. These labels are either `true' or `false', indicating whether the content contains a propagandistic technique. As observed in our analysis, there is a significant skew in the label distribution. As shown in Table \ref{class-distribution}, only 21\% of the data is labeled as `false,' while the remaining 79\% carries a `true' label. This imbalance in classes could introduce challenges during the training phase. Furthermore, we found that 64.9\% of the data originates from paragraphs, while the remaining 35.1\% is sourced from tweets.

\begin{table*}[h]
\centering
\setlength{\tabcolsep}{3pt}
\scalebox{0.95}{
\begin{tabular}{l|r|r|r|r}
\toprule
 & \multicolumn{2}{c|}{\textbf{Task 1A}} & \multicolumn{2}{c}{\textbf{Task 2A}} \\
\midrule
 & \textbf{Prop} & \textbf{Non-Prop} & \textbf{Disinfo} & \textbf{No-Disinfo} \\
\hline
Train & 1,918 (79\%)& 509\ (21\%)& 2,656\ (19.8\%)& 11,491\ (81.2\%)\\
Dev & 202\ (78\%)& 57\ (22\%)& 397\ (18.8\%)& 1,718\ (81.2\%)\\
Test & 331\ (65.8\%)& 172\ (34.2\%)& 876\ (23.8\%)& 2,853\ (76.2\%)\\
\textbf{Total} &     2451&    733&      3929&  15062\\
\bottomrule
\end{tabular}
}
\caption{Class label distribution for task 1A and 2A. Prop. -- Contains propagandistic technique; Non-Prop -- does not contain any propagandistic technique.}
\label{class-distribution}
\end{table*}


\paragraph{Subtask 2A:} Given a tweet, categorize whether it is disinformative. This is a binary classification task. 

The data format for Subtask 2A is identical to that of Subtask 1A. Similar to Subtask 1A, this subtask also shows a skewed label distribution. Specifically, only 18.8\% of the data is tagged as \textbf{disinfo}, while the remaining 79\% carries the \textbf{no-disinfo} tag, as can be seen in Table \ref{class-distribution}. This imbalance in class distribution could present challenges during the model training process.




For our experiments, we used the same training, development, and test datasets as provided by the organizers. Details on the data distribution can be found in Table \ref{class-distribution}.

\paragraph{Evaluation Measures:} The official evaluation metric for Subtask A is Micro-F1, while for Subtask B, it is Macro-F1. 

\section{Methodology}
\label{sec:methodology}


\subsection{Pre-trained Models}
Given that large-scale pre-trained Transformer models have achieved state-of-the-art performance for several NLP tasks. Therefore, as deep learning algorithms, we used deep contextualized text representations based on such pre-trained transformer models. We used AraBERT \citep{baly2020arabert}, MarBERT~\cite{abdul-mageed-etal-2021-arbert} and Qarib~\cite{abdelali2021pre} due to their promising performance in other Arabic NLP tasks. 

Consequently, text preprocessing was done using the AraBERT preprocessor with the default configuration. Hyperparameters were tuned and op-timized through the use of randomized grid search. The chosen configuration for the task involved a maximum tokenization length of 128, a batch size of 16, running for a total of 3 epochs during training, with a learning rate set at 4e-5, and utilizing the AdamW optimizer. As a loss function, we used cross-entropy loss:

$$
\text{CrossEntropyLoss} = -\sum_{i=1}^{N} \sum_{j=1}^{C} y_{ij} \cdot \log(p_{ij})
$$
where, $N$ is the number of samples, $C$ is the number of classes, $y_{ij}$ is the ground truth label (1 if the sample $i$ belongs to class $j$, 0 otherwise), and $p_{ij}$ is the predicted probability of sample $i$ belonging to class $j$. 

After closely examining the weights in the cross-entropy loss function, we chose to assign four times the weight to the `false' tag compared to the `true' tag, resulting in a weight array of [1.0, 4.0] for the cross-entropy loss.

\begin{figure}
     \centering
     \includegraphics[width=1\linewidth]{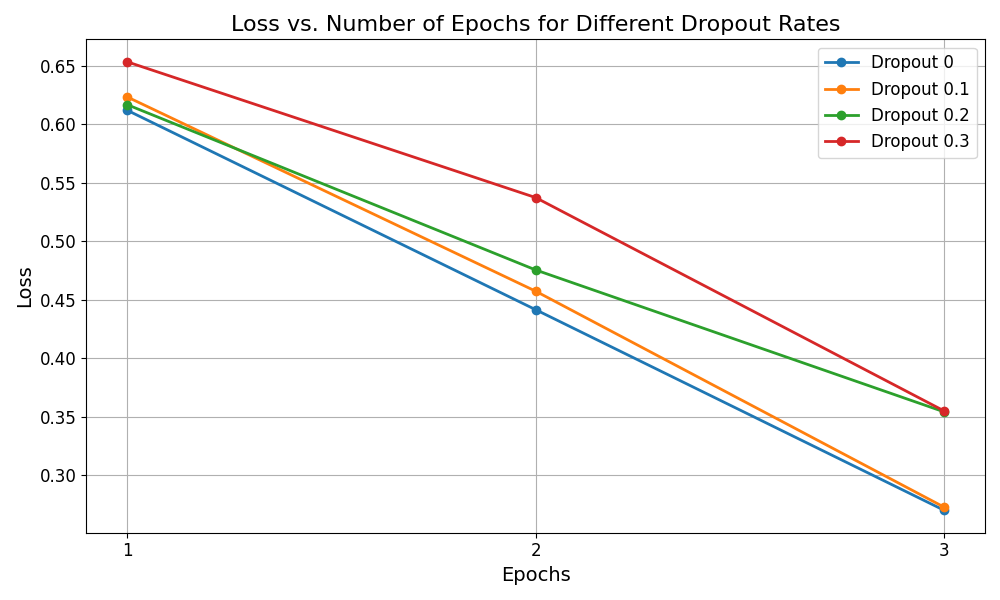}
     \caption{Loss per epoch with different dropout rate.}
     \label{fig:dropout}
 \end{figure}

Additionally, we observed that the dataset is highly imbalanced. Incorporating a dropout layer improved the model's performance. To optimize this, we experimented with varying dropout rates and monitored the corresponding loss across different epochs, as illustrated in Figure \ref{fig:dropout}.

Surprisingly, the models with lower dropout rates, which exhibited lower loss in the final epoch, performed worse than those with slightly higher dropout rates. We suspect that the models may have overfitted when using lower dropout rates, resulting in subpar performance on the test set.



\subsection{Large Language Models (LLMs)}
For the LLMs, we investigate their performance in both in-context zero-shot and few-shot learning settings. This involves prompting and post-processing the output to extract the expected content. We utilized GPT-4 \cite{openai2023gpt} in both zero-shot and few-shot settings for both subtasks. To ensure reproducibility, we set the temperature to zero for all settings. Note that for GPT-4, we used version 0314, which was released in June 2023. Our choice of this model was based on its accessibility. For the experiments, we employed the LLMeBench framework \cite{dalvi2023llmebench}, following the prompts and instructions previously studied for Arabic in \cite{abdelali2023benchmarking}. 

\begin{table}[ht]
\centering
\setlength{\tabcolsep}{4pt}
\scalebox{0.88}{
\begin{tabular}{lrrrrr}
\toprule
\multirow{2}{*}{\textbf{Model}} & \multirow{2}{*}{\textbf{Dropout}} & \multicolumn{2}{c}{\textbf{Micro F1}} & \multicolumn{2}{c}{\textbf{Macro F1}} \\
\cline{3-6}
& & \textbf{Dev} & \textbf{Test} & \textbf{Dev} & \textbf{Test} \\
\midrule
\textbf{Submission} &  & & 0.740& & 0.693\\ \midrule
\multirow{4}{*}{\textbf{AraBERT}} & 0 & 0.656& 0.625& 0.723& 0.712\\
& 0.1 & 0.772& 0.704& 0.725& 0.714\\
& 0.2 & 0.772& 0.692& 0.739& 0.740\\
& 0.3 & n/a & n/a & 0.743& 0.713\\
\midrule
\multirow{4}{*}{\textbf{MarBERT}}& 0 & 0.810& \textbf{0.756}& 0.707& 0.696\\
& 0.1 & 0.841& 0.731& 0.745& 0.718\\
& 0.2 & 0.818& 0.746& 0.769& 0.731\\
& 0.3 & n/a & n/a & 0.737& 0.708 \\ 
\bottomrule
\end{tabular}
}
\caption{Results with different dropout rates and submitted system for subtask 1A. n/a refers to the number was not ready at time of preparing the paper.}
\label{tab:model-performance}
\end{table}

\begin{table}[htbp]
\centering
\setlength{\tabcolsep}{4pt}
\scalebox{0.90}{
\begin{tabular}{@{}lccc@{}}
\toprule
\multirow{3}{*}{\textbf{Model}} & \multirow{3}{*}{\textbf{Dropout}} & \multicolumn{2}{c}{\textbf{Test}} \\
\cmidrule(lr){3-4}
& & \textbf{Micro} & \textbf{Macro} \\
& & \textbf{F1} & \textbf{F1} \\
\midrule
Submission &  0.2&  0.893&  0.845\\ \midrule
Qarib & 0 & 0.889 & 0.822 \\
& 0.1 & 0.898 & 0.844 \\
& 0.2 & 0.903 & \textbf{0.869} \\
& 0.3 & 0.897 & 0.849 \\
\midrule
MarBERT & 0.1 & 0.898 & 0.843 \\
& 0.2 & 0.898 & 0.846 \\
& 0.3 & 0.899 & 0.849 \\
\midrule
AraBERT & 0 & 0.802 & 0.794 \\
& 0.1 & 0.846 & 0.813 \\
& 0.2 & 0.893 & 0.846 \\
\bottomrule
\end{tabular}
}
\caption{Model performance with different dropout rates and submitted system for subtask 2A (disinformative vs. not-disinformative).}
\label{tab:2A-performance}
\end{table}

\section{Results and Discussion}
\label{sec:results_and_discussion}

\subsection{Subtask 1A}
For this shared task, we were given a dataset containing 504 text entries. We employed the model described in the previous section to predict various labels for each tweet. The final results released by the task organizers indicated that our model achieved a Micro F1 of 0.740 and a Macro F1 of 0.693. In Table \ref{tab:model-performance}, we present the performance metrics for our submitted system, comparing them with other models and various dropout rates.

Through our discovery, we realize that MarBERT performed extremely well compared to Arabert. This is expected as MarBERT is trained on tweets, which is very similar to the data provided. Nevertheless, we found it even more surprising that MarBERT's performance dropped after applying the dropout layer. This potentially indicates that the model might be undertrained and we might need to run a few more epochs.

\subsection{Subtask 2A}
For this shared task, we are provided with 3729 entries of text. The model described in the previous section was used to predict various labels for each tweet.  The final results released by the task organizers have shown that the model that we have
scored 0.7396 in Micro F1 and 0.74 in Macro F1. In Table \ref{tab:2A-performance} we have displayed some of our attempts, and after more experiments we are able to achieve higher result.

We noticed that in task2A that qarib outperformed MarBERT, despite both trained using a variety of tweets.  This could be the result of better/bigger training set or the result of longer training duration. To discover why, further investigation and experimentation have to be made.

In Table \ref{tab:gpt4_results}, we report the results on the test sets for both tasks with zero and 5-shots learning using GPT-4. It appears that the performances are significantly lower than fine-tuned models. We see an improvement with 5-shots, which was also observed in prior studies \cite{abdelali2023benchmarking}. However, such performances are still lower than fine-tuned models. Further studies are required to understand their capabilities as prompt engineering is the key factor to achive a desired results with LLMs. 
\begin{table}[]
\centering
\setlength{\tabcolsep}{4pt}
\scalebox{0.85}{
\begin{tabular}{@{}llrr@{}}
\toprule
\multicolumn{1}{c}{} & \multicolumn{1}{c}{\textbf{Shot}} & \multicolumn{1}{c}{\textbf{Micro F1}} & \multicolumn{1}{c}{\textbf{Macro F1}} \\ \midrule
Task 1A & 0-shot & 0.600 & 0.600 \\
 & 5-shot & 0.614 & 0.614 \\
Task 2A & 0-shot & 0.759 & 0.707  \\
 & 5-shot &  0.852  & 0.804 \\ \bottomrule
\end{tabular}%
}
\caption{Results on the test set with zero- and few-shot learning using GPT-4.}
\label{tab:gpt4_results}
\end{table}

\section{Conclusion and Future work}
\label{sec:conclusion}
In this paper, we report on our participation in the ArAIEval 2023 shared task, which focuses on propaganda and disinformation detection. We experimented with various transformer-based models and fine-tuned them for our specific tasks. Despite challenges such as imbalanced data, we optimized our models and achieved commendable results. Our submitted system ranked 9th and 10th in subtasks 1A and 2A, respectively, on the leaderboard. 
In the future, our research will take advantage of the latest Large Language Models (LLMs) such as Llama, Alpaca, Bloom and more. We plan to do more experiment with data augmentation. 

\section*{Limitations}
Our study primarily focused on fine-tuned transformer-based models and zero-shot and few-shot learning with GPT-4. Given that the dataset is heavily skewed towards certain classes, our study did not address these aspects. However, this will be the focus of a future study.   

\bibliography{bib/bibliography,bib/propaganda,bib/disinfo}

\begin{thebibliography}{30}
\expandafter\ifx\csname natexlab\endcsname\relax\def\natexlab#1{#1}\fi

\bibitem[{Abdelali et~al.(2021)Abdelali, Hassan, Mubarak, Darwish, and
  Samih}]{abdelali2021pre}
Ahmed Abdelali, Sabit Hassan, Hamdy Mubarak, Kareem Darwish, and Younes Samih.
  2021.
\newblock Pre-training bert on arabic tweets: Practical considerations.
\newblock \emph{arXiv preprint arXiv:2102.10684}.

\bibitem[{Abdelali et~al.(2023)Abdelali, Mubarak, Chowdhury, Hasanain, Mousi,
  Boughorbel, Kheir, Izham, Dalvi, Hawasly, Nazar, Elshahawy, Ali, Durrani,
  Milic-Frayling, and Alam}]{abdelali2023benchmarking}
Ahmed Abdelali, Hamdy Mubarak, Shammur~Absar Chowdhury, Maram Hasanain, Basel
  Mousi, Sabri Boughorbel, Yassine~El Kheir, Daniel Izham, Fahim Dalvi, Majd
  Hawasly, Nizi Nazar, Yousseif Elshahawy, Ahmed Ali, Nadir Durrani, Natasa
  Milic-Frayling, and Firoj Alam. 2023.
\newblock \href {http://arxiv.org/abs/2305.14982} {Benchmarking arabic ai with
  large language models}.

\bibitem[{Abdul-Mageed et~al.(2021)Abdul-Mageed, Elmadany, and
  Nagoudi}]{abdul-mageed-etal-2021-arbert}
Muhammad Abdul-Mageed, AbdelRahim Elmadany, and El~Moatez~Billah Nagoudi. 2021.
\newblock \href {https://doi.org/10.18653/v1/2021.acl-long.551} {{ARBERT} {\&}
  {MARBERT}: Deep bidirectional transformers for {A}rabic}.
\newblock In \emph{Proceedings of the 59th Annual Meeting of the Association
  for Computational Linguistics and the 11th International Joint Conference on
  Natural Language Processing (Volume 1: Long Papers)}, pages 7088--7105,
  Online. Association for Computational Linguistics.

\bibitem[{Alam et~al.(2022{\natexlab{a}})Alam, Cresci, Chakraborty, Silvestri,
  Dimitrov, Martino, Shaar, Firooz, and Nakov}]{alam-etal-2022-survey}
Firoj Alam, Stefano Cresci, Tanmoy Chakraborty, Fabrizio Silvestri, Dimiter
  Dimitrov, Giovanni Da~San Martino, Shaden Shaar, Hamed Firooz, and Preslav
  Nakov. 2022{\natexlab{a}}.
\newblock \href {https://aclanthology.org/2022.coling-1.576} {A survey on
  multimodal disinformation detection}.
\newblock In \emph{Proceedings of the 29th International Conference on
  Computational Linguistics}, pages 6625--6643, Gyeongju, Republic of Korea.
  International Committee on Computational Linguistics.

\bibitem[{Alam et~al.(2022{\natexlab{b}})Alam, Mubarak, Zaghouani, Nakov, and
  Da~San~Martino}]{propaganda-detection:WANLP2022-overview}
Firoj Alam, Hamdy Mubarak, Wajdi Zaghouani, Preslav Nakov, and Giovanni
  Da~San~Martino. 2022{\natexlab{b}}.
\newblock Overview of the {WANLP} 2022 shared task on propaganda detection in
  {A}rabic.
\newblock In \emph{Proceedings of the Seventh Arabic Natural Language
  Processing Workshop}, Abu Dhabi, UAE. Association for Computational
  Linguistics.

\bibitem[{Antoun et~al.(2020)Antoun, Baly, and Hajj}]{baly2020arabert}
Wissam Antoun, Fady Baly, and Hazem Hajj. 2020.
\newblock {AraBERT}: Transformer-based model for {Arabic} language
  understanding.
\newblock In \emph{Proceedings of the 4th Workshop on Open-Source Arabic
  Corpora and Processing Tools, with a Shared Task on Offensive Language
  Detection}, pages 9--15.

\bibitem[{Barr{\'o}n-Cedeno et~al.(2019)Barr{\'o}n-Cedeno, Jaradat,
  Da~San~Martino, and Nakov}]{BARRONCEDENO20191849}
Alberto Barr{\'o}n-Cedeno, Israa Jaradat, Giovanni Da~San~Martino, and Preslav
  Nakov. 2019.
\newblock Proppy: Organizing the news based on their propagandistic content.
\newblock \emph{In IPM}, 56(5):1849--1864.

\bibitem[{Boulouard et~al.(2022)Boulouard, Ouaissa, Ouaissa, Krichen, Almutiq,
  and Karim}]{Boulouard22}
Zakaria Boulouard, Mariya Ouaissa, Mariyam Ouaissa, Moez Krichen, Mutiq
  Almutiq, and Gasmi Karim. 2022.
\newblock \href {https://doi.org/10.3390/app122412823} {Detecting hateful and
  offensive speech in arabic social media using transfer learning}.
\newblock \emph{Applied Sciences}, 12:12823.

\bibitem[{Da~San~Martino et~al.(2020)Da~San~Martino, Barr{\'o}n-Cedeno,
  Wachsmuth, Petrov, and Nakov}]{da2020semeval}
Giovanni Da~San~Martino, Alberto Barr{\'o}n-Cedeno, Henning Wachsmuth,
  Rostislav Petrov, and Preslav Nakov. 2020.
\newblock {SemEval}-2020 task 11: Detection of propaganda techniques in news
  articles.
\newblock In \emph{Proceedings of the 14th Workshop on Semantic Evaluation},
  SemEval~'20, pages 1377--1414.

\bibitem[{Da~San~Martino et~al.(2019)Da~San~Martino, Yu, Barr{\'o}n-Cede{\~n}o,
  Petrov, and Nakov}]{EMNLP19DaSanMartino}
Giovanni Da~San~Martino, Seunghak Yu, Alberto Barr{\'o}n-Cede{\~n}o, Rostislav
  Petrov, and Preslav Nakov. 2019.
\newblock \href {https://doi.org/10.18653/v1/D19-1565} {Fine-grained analysis
  of propaganda in news article}.
\newblock In \emph{EMNLP-IJCNLP}, pages 5636--5646.

\bibitem[{Dalvi et~al.(2023)Dalvi, Hasanain, Boughorbel, Mousi, Abdaljalil,
  Nazar, Abdelali, Chowdhury, Mubarak, Ali, Hawasly, Durrani, and
  Alam}]{dalvi2023llmebench}
Fahim Dalvi, Maram Hasanain, Sabri Boughorbel, Basel Mousi, Samir Abdaljalil,
  Nizi Nazar, Ahmed Abdelali, Shammur~Absar Chowdhury, Hamdy Mubarak, Ahmed
  Ali, Majd Hawasly, Nadir Durrani, and Firoj Alam. 2023.
\newblock \href {http://arxiv.org/abs/2308.04945} {{LLMeBench}: A flexible
  framework for accelerating llms benchmarking}.
\newblock \emph{arXiv:2308.04945}.

\bibitem[{Demilie and Salau(2022)}]{Demilie22}
W.B. Demilie and A.O. Salau. 2022.
\newblock \href {https://doi.org/10.1186/s40537-022-00619-x} {Detection of fake
  news and hate speech for ethiopian languages: a systematic review of the
  approaches.}
\newblock \emph{J Big Data}.

\bibitem[{Devlin et~al.(2018)Devlin, Chang, Lee, and
  Toutanova}]{devlin2018bert}
Jacob Devlin, Ming-Wei Chang, Kenton Lee, and Kristina Toutanova. 2018.
\newblock {BERT}: Pre-training of deep bidirectional transformers for language
  understanding.
\newblock \emph{arXiv preprint arXiv:1810.04805}.

\bibitem[{Dimitrov et~al.(2021{\natexlab{a}})Dimitrov, Bin~Ali, Shaar, Alam,
  Silvestri, Firooz, Nakov, and Da~San~Martino}]{dimitrov2021detecting}
Dimitar Dimitrov, Bishr Bin~Ali, Shaden Shaar, Firoj Alam, Fabrizio Silvestri,
  Hamed Firooz, Preslav Nakov, and Giovanni Da~San~Martino. 2021{\natexlab{a}}.
\newblock \href {https://doi.org/10.18653/v1/2021.acl-long.516} {Detecting
  propaganda techniques in memes}.
\newblock In \emph{ACL-IJCNLP}, ACL-IJCNLP~'21, pages 6603--6617, Online.
  Association for Computational Linguistics.

\bibitem[{Dimitrov et~al.(2021{\natexlab{b}})Dimitrov, Bin~Ali, Shaar, Alam,
  Silvestri, Firooz, Nakov, and Da~San~Martino}]{SemEval2021-6-Dimitrov}
Dimitar Dimitrov, Bishr Bin~Ali, Shaden Shaar, Firoj Alam, Fabrizio Silvestri,
  Hamed Firooz, Preslav Nakov, and Giovanni Da~San~Martino. 2021{\natexlab{b}}.
\newblock {Task 6 at SemEval-2021}: Detection of persuasion techniques in texts
  and images.
\newblock In \emph{SemEval}.

\bibitem[{Fortuna and Nunes(2018)}]{fortuna2018survey}
Paula Fortuna and S{\'e}rgio Nunes. 2018.
\newblock A survey on automatic detection of hate speech in text.
\newblock \emph{CSUR}, 51(4):1--30.

\bibitem[{Hasanain et~al.(2023{\natexlab{a}})Hasanain, Alam, Mubarak,
  Abdaljalil, Zaghouani, Nakov, Da~San~Martino, and
  Freihat}]{araieval:arabicnlp2023-overview}
Maram Hasanain, Firoj Alam, Hamdy Mubarak, Samir Abdaljalil, Wajdi Zaghouani,
  Preslav Nakov, Giovanni Da~San~Martino, and Abed~Alhakim Freihat.
  2023{\natexlab{a}}.
\newblock {ArAIEval Shared Task: Persuasion Techniques and Disinformation
  Detection in Arabic Text}.
\newblock In \emph{{Proceedings of the First Arabic Natural Language Processing
  Conference (ArabicNLP 2023)}}, Singapore. Association for Computational
  Linguistics.

\bibitem[{Hasanain et~al.(2023{\natexlab{b}})Hasanain, El-Shangiti, Nandi,
  Nakov, and Alam}]{hasanain-etal-2023-qcri}
Maram Hasanain, Ahmed El-Shangiti, Rabindra~Nath Nandi, Preslav Nakov, and
  Firoj Alam. 2023{\natexlab{b}}.
\newblock \href {https://doi.org/10.18653/v1/2023.semeval-1.172} {{QCRI} at
  {S}em{E}val-2023 task 3: News genre, framing and persuasion techniques
  detection using multilingual models}.
\newblock In \emph{Proceedings of the 17th International Workshop on Semantic
  Evaluation (SemEval-2023)}, pages 1237--1244, Toronto, Canada. Association
  for Computational Linguistics.

\bibitem[{Li et~al.(2016)Li, Gao, Meng, Li, Su, Zhao, Fan, and
  Han}]{li2016survey}
Yaliang Li, Jing Gao, Chuishi Meng, Qi~Li, Lu~Su, Bo~Zhao, Wei Fan, and Jiawei
  Han. 2016.
\newblock A survey on truth discovery.
\newblock \emph{ACM Sigkdd Explorations Newsletter}, 17(2):1--16.

\bibitem[{Miller(1939)}]{Miller}
Clyde~R. Miller. 1939.
\newblock {The Techniques of Propaganda}.
\newblock pages 27--29.

\bibitem[{Mubarak et~al.(2023)Mubarak, Abdaljalil, Nassar, and
  Alam}]{mubarak2023detecting}
Hamdy Mubarak, Samir Abdaljalil, Azza Nassar, and Firoj Alam. 2023.
\newblock Detecting and reasoning of deleted tweets before they are posted.
\newblock \emph{arXiv preprint arXiv:2305.04927}.

\bibitem[{Mubarak et~al.(2020)Mubarak, Darwish, Magdy, Elsayed, and
  Al-Khalifa}]{mubarak2020overview}
Hamdy Mubarak, Kareem Darwish, Walid Magdy, Tamer Elsayed, and Hend Al-Khalifa.
  2020.
\newblock Overview of osact4 arabic offensive language detection shared task.
\newblock In \emph{Proceedings of the 4th Workshop on open-source arabic
  corpora and processing tools, with a shared task on offensive language
  detection}, pages 48--52.

\bibitem[{OpenAI(2023)}]{openai2023gpt}
R~OpenAI. 2023.
\newblock Gpt-4 technical report.
\newblock \emph{arXiv}, pages 2303--08774.

\bibitem[{Oshikawa et~al.(2018)Oshikawa, Qian, and Wang}]{oshikawa2018survey}
Ray Oshikawa, Jing Qian, and William~Yang Wang. 2018.
\newblock A survey on natural language processing for fake news detection.
\newblock \emph{arXiv preprint arXiv:1811.00770}.

\bibitem[{Oshikawa et~al.(2020)Oshikawa, Qian, and
  Wang}]{oshikawa-etal-2020-survey}
Ray Oshikawa, Jing Qian, and William~Yang Wang. 2020.
\newblock A survey on natural language processing for fake news detection.
\newblock In \emph{Proceedings of the 12th Language Resources and Evaluation
  Conference}, LREC~'20, pages 6086--6093.

\bibitem[{Rashkin et~al.(2017)Rashkin, Choi, Jang, Volkova, and
  Choi}]{rashkin-EtAl:2017:EMNLP2017}
Hannah Rashkin, Eunsol Choi, Jin~Yea Jang, Svitlana Volkova, and Yejin Choi.
  2017.
\newblock \href {https://doi.org/10.18653/v1/D17-1317} {Truth of varying
  shades: {A}nalyzing language in fake news and political fact-checking}.
\newblock In \emph{Proceedings of the 2017 Conference on Empirical Methods in
  Natural Language Processing}, pages 2931--2937. Association for Computational
  Linguistics.

\bibitem[{Sharma et~al.(2022)Sharma, Alam, Akhtar, Dimitrov, Da~San~Martino,
  Firooz, Halevy, Silvestri, Nakov, and Chakraborty}]{ijcai2022p781}
Shivam Sharma, Firoj Alam, Md.~Shad Akhtar, Dimitar Dimitrov, Giovanni
  Da~San~Martino, Hamed Firooz, Alon Halevy, Fabrizio Silvestri, Preslav Nakov,
  and Tanmoy Chakraborty. 2022.
\newblock \href {https://doi.org/10.24963/ijcai.2022/781} {Detecting and
  understanding harmful memes: A survey}.
\newblock In \emph{Proceedings of the Thirty-First International Joint
  Conference on Artificial Intelligence, {IJCAI-22}}, pages 5597--5606.
  International Joint Conferences on Artificial Intelligence Organization.
\newblock Survey Track.

\bibitem[{Zampieri et~al.(2019)Zampieri, Malmasi, Nakov, Rosenthal, Farra, and
  Kumar}]{Offensive:Type:Target:2019}
Marcos Zampieri, Shervin Malmasi, Preslav Nakov, Sara Rosenthal, Noura Farra,
  and Ritesh Kumar. 2019.
\newblock Predicting the type and target of offensive posts in social media.
\newblock In \emph{NAACL-HLT}, pages 1415--1420.

\bibitem[{Zampieri et~al.(2020)Zampieri, Nakov, Rosenthal, Atanasova,
  Karadzhov, Mubarak, Derczynski, Pitenis, and
  {\c{C}}{\"o}ltekin}]{zampieri2020semeval}
Marcos Zampieri, Preslav Nakov, Sara Rosenthal, Pepa Atanasova, Georgi
  Karadzhov, Hamdy Mubarak, Leon Derczynski, Zeses Pitenis, and
  {\c{C}}a{\u{g}}r{\i} {\c{C}}{\"o}ltekin. 2020.
\newblock {S}em{E}val-2020 task 12: Multilingual offensive language
  identification in social media ({O}ffens{E}val 2020).
\newblock In \emph{SemEval}, pages 1425--1447.

\bibitem[{Zhou et~al.(2016)Zhou, Wang, and Chen}]{Zhou2016}
Lu~Zhou, Wenbo Wang, and Keke Chen. 2016.
\newblock \href {https://doi.org/10.1145/2872427.2883052} {Tweet properly:
  Analyzing deleted tweets to understand and identify regrettable ones}.
\newblock In \emph{Proceedings of the 25th International Conference on World
  Wide Web}, WWW '16, page 603–612, Republic and Canton of Geneva, CHE.
  International World Wide Web Conferences Steering Committee.

\end{thebibliography}
\bibliographystyle{acl_natbib}




\end{document}